\def\year{2019}\relax

\documentclass[letterpaper]{article} 
\usepackage{aaai19}  
\usepackage{hyperref}
\usepackage{times}
\usepackage{amsfonts}
\usepackage{amsmath}
\usepackage{graphicx}
\usepackage{url}
\usepackage{booktabs} 
\usepackage{nicefrac}
\usepackage{subfigure}
\usepackage{multirow}
\newcommand{\citet}[1]{\citeauthor{#1}}

\title{Stacked Pooling: Improving Crowd Counting by Boosting Scale Invariance}

\author{Siyu Huang$^{1}$, Xi Li$^{1}$, Zhi-Qi Cheng$^{2}$, Zhongfei Zhang$^1$, Alexander Hauptmann$^3$ \\
$^1$Zhejiang University, $^2$Southwest Jiaotong University, $^3$Carnegie Mellon University \\
\tt{\{siyuhuang,xilizju,zhongfei\}@zju.edu.cn, zhiqicheng@gmail.com, alex@cs.cmu.edu} \\}

\begin{document}
\maketitle
\begin{abstract}
In this work, we explore the cross-scale similarity in crowd counting scenario, in which the regions of different scales often exhibit high visual similarity. This feature is universal both within an image and across different images, indicating the importance of scale invariance of a crowd counting model. Motivated by this, in this paper we propose simple but effective variants of pooling module, i.e., multi-kernel pooling and stacked pooling, to boost the scale invariance of convolutional neural networks (CNNs), benefiting much the crowd density estimation and counting. Specifically, the multi-kernel pooling comprises of pooling kernels with multiple receptive fields to capture the responses at multi-scale local ranges. The stacked pooling is an equivalent form of multi-kernel pooling, while, it reduces considerable computing cost. Our proposed pooling modules do not introduce extra parameters into model and can easily take place of the vanilla pooling layer in implementation. In empirical study on two benchmark crowd counting datasets, the stacked pooling beats the vanilla pooling layer in most cases.
\end{abstract}

\section{Introduction}
Crowd counting has been widely studied for decades of years because of a great many practical demands such as public safety and city planning. While, crowd counting still remains challenging and researchers seek to address it by focusing on aspects of severe occlusions, perspective distortions, and diverse crowd distributions. 

\begin{figure}[t]
\centering
\includegraphics[width=1\linewidth]{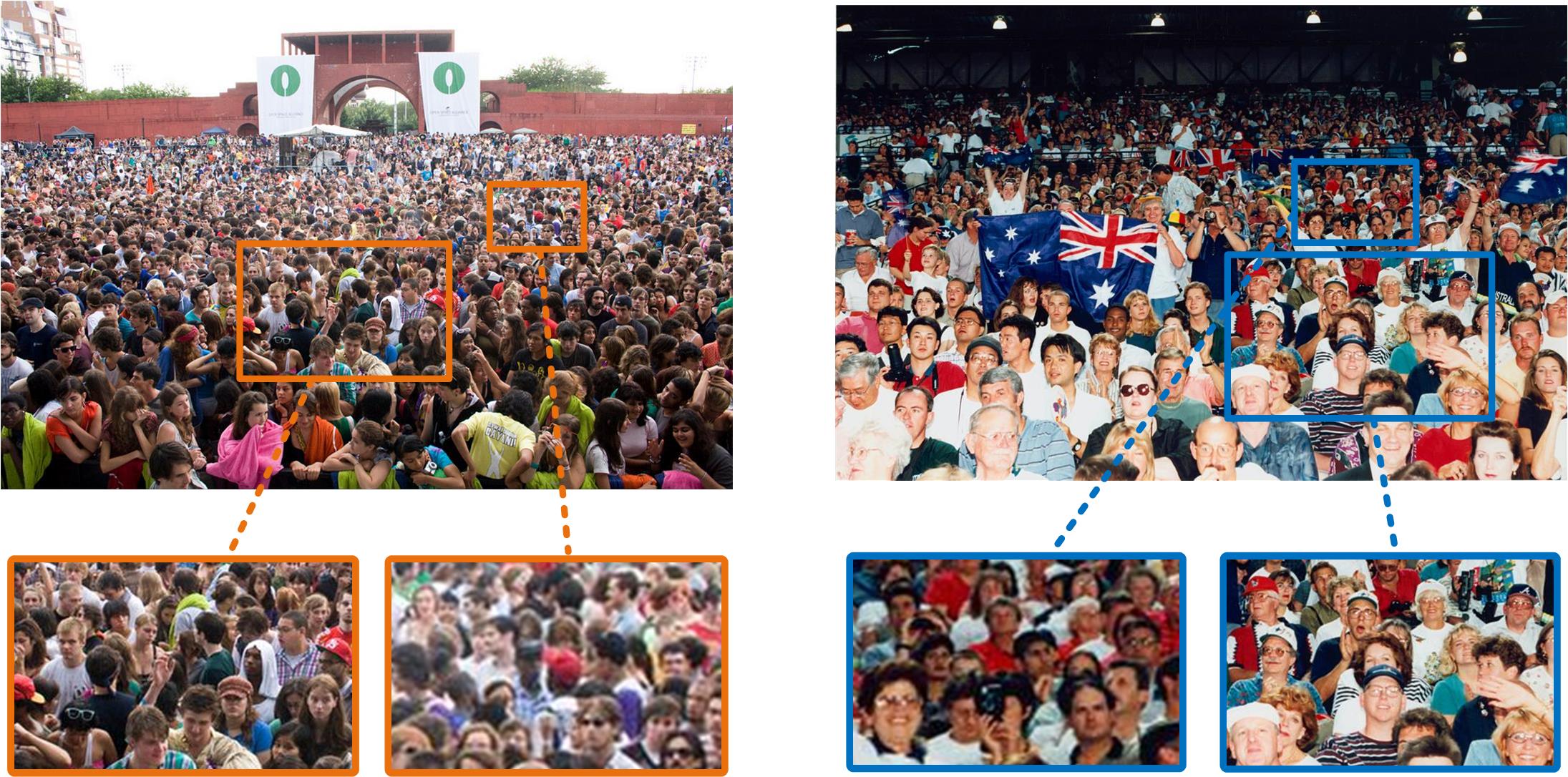}
\vspace{1pt}
\caption{\textbf{In crowd images, regions of different scales exhibit high visual similarity if we resize them to certain sizes.} This feature is common for regions within an image and also for regions among different images. It indicates the importance of scale invariance in crowd counting.} 
\label{example_similarity}
\end{figure}

In this work, we explore an important feature in crowd counting scenario, i.e., \emph{cross-scale visual similarity}. Fig. \ref{example_similarity} shows two examples of the cross-scale visual similarity within crowd images. In each image, it is not difficult to find two regions which are visually similar when they are resized to the same scale. The cross-scale visual similarity is quite universal in crowd counting, not only within an individual image, but also among different images of various scenes. In contrast, this feature is not typical for the natural images such as Cifar-10 \cite{krizhevsky2009learning} and ImageNet \cite{deng2009imagenet}. Therefore, the vision model designed for crowd counting especially requires the capability of scale invariance.

A common solution for augmenting scale invariance of convolutional neural networks (CNNs) would be to make CNNs larger \cite{krizhevsky2012imagenet,simonyan2014very,huang2017densely} and deeper \cite{szegedy2015going,he2016deep} by introducing more learnable parameters to improve their representation performances. Another solution is to manually build branches in CNNs for visual concepts of different scales \cite{xu2014scale,cai2016unified}. Specifically in crowd counting, researchers explore various variants \cite{zeng2017multi,sam2017switching} of multi-sized convolutions \cite{zhang2016single} to deal with the scale variation in people size. 

\begin{figure}[t]
\centering
\includegraphics[width=1\linewidth]{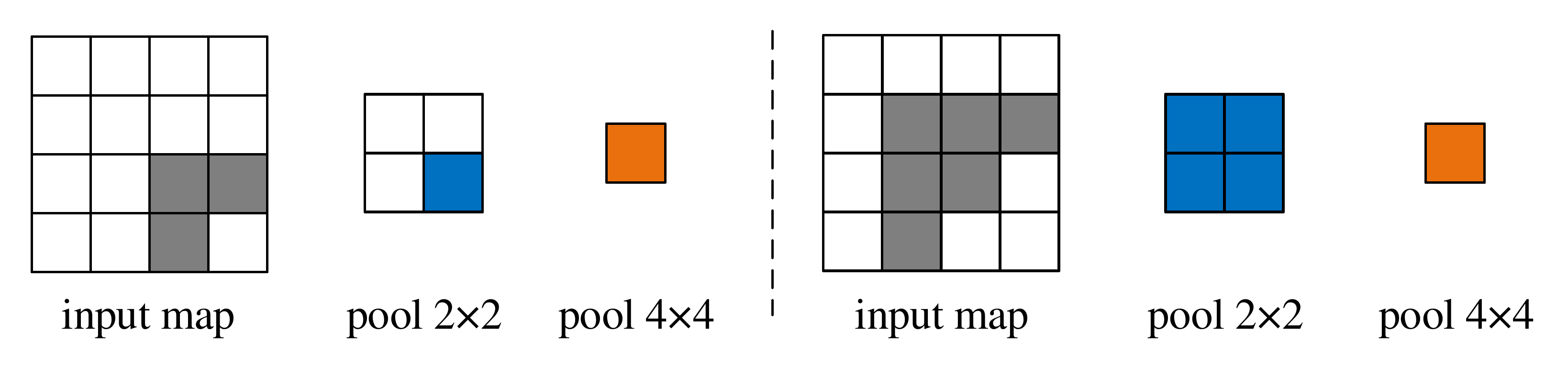}
\caption{An intuitive illustration of the scale invariance brought by a larger pooling kernel.} 
\label{invariance_by_pooling}
\end{figure}

Different from these approaches, we focus on the pooling module to boost the scale invariance of CNNs. As studied by existing literature \cite{huang2007unsupervised,boureau2010theoretical,scherer2010evaluation}, the scale invariance of CNNs is in general brought by the pooling layer. However, it is evident that the conventional pooling can only deal with slight scale change \cite{gong2014multi}, thus cannot well cope with the significant scale variation in crowd counting scenarios, e.g., the examples shown in Fig.\ref{example_similarity}. Motivated by this, we propose to employ a larger pooling range to adapt the network to such a severe scale variation. Fig. \ref{invariance_by_pooling} illustrates how a larger pooling range enables an invariance when the input goes through a scale variation. The feature map after 2$\times$2 max-pooling changes. While, the feature map after 4$\times$4 max-pooling remains unchanged, presenting a scale invariance.

In this paper, we propose simple but effective variants of pooling module, i.e., multi-kernel pooling and stacked pooling, to boost the scale invariance of CNNs. Specifically, the multi-kernel pooling comprises of pooling kernels with multiple receptive fields to capture the responses at multi-scale local ranges, then, concatenating the feature maps together to its successive layer. Technically, the larger pooling kernels can provide a wider range of scale invariance for CNNs, while the fine-grained information is also preserved by smaller pooling kernels. The stacked pooling is an equivalent form of multi-kernel pooling by stacking smaller pooling kernels. It further reduces the computing cost of multi-kernel pooling. In practice, our proposed pooling modules have the following advantages:
\begin{itemize}
\item \emph{Non-parametric:} They do not introduce extra parameters and hyper-parameters into the model, ensuring the model efficiency and preventing the overfitting in learning.
\item \emph{Simple and flexible:} They are succinct and very easy to implement. They can take place of the vanilla pooling layer at any time when need be.
\end{itemize}

In empirical study, the stacked pooling shows favorable performance in comparison with the vanilla pooling. It beats the vanilla pooling in most experiments on two benchmark crowd counting datasets. In addition, insight studies about pooling kernel sizes and their impact on the scale variance of CNNs further reveal the effectiveness of stacked pooling. 

We summarize the contributions of this paper as follow:
\begin{itemize}
\item We explore the cross-scale visual similarity in crowd images to promote the study on the scale invariance of crowd counting models.
\item We propose multi-kernel pooling and stacked pooling which are simple and flexible variants of vanilla pooling for boosting the scale invariance of CNNs and improving crowd counting performances.  
\item We empirically demonstrate the effectiveness of proposed pooling modules and take insight into their impact on the invariance of CNNs when facing scale variation.
\end{itemize}

\section{Related Work}
\subsection{Deep crowd counting} The deep CNNs are currently the state-of-the-art approach \cite{sindagi2017cnn,sindagi2017generating,liu2018decidenet,babu2018divide,li2018structured,huang2018body}for crowd density estimation and crowd counting due to their powerful visual representation ability.

Specifically to deal with the large scale variation in people size, researchers mainly focused on the improvement of convolution units in recent years \cite{zeng2017multi,zhang2018crowd,li2018csrnet}. For a typical example, the Multi-Column CNN \cite{zhang2016single} and Switching CNN \cite{sam2017switching} exploited multi-sized convolutional kernels to adapt CNNs to people of different sizes. Another popular approach is to transform the scale of the feature map to adapt feature itself to scale variation. For instance, the Hydra CNN \cite{onoro2016towards} adopted a pyramid of multi-scale image patches as input such that each branch of CNN learns the feature representation for a particular scale of the pyramid. \citeauthor{shen2018crowd} proposed a scale-consistency regularization constraint to integrate large-scale and small-scale images.

Different from all of these approaches, this work focuses on the pooling layer, as it is generally assumed that the pooling layer enables the scale invariance of CNNs. Motivated by the significant scale variation in crowd counting, we propose the multi-kernel pooling to take place of the vanilla pooling module, aiming at more scale-invariant CNNs.

\subsection{Variants of pooling}
Various variants of pooling have been proposed by the computer vision community \cite{boureau2011ask,yoo2015multi}. For instance, the well-known L2 pooling \cite{aapo2009natural,ngiam2010tiled} is proposed towards the complex invariances of CNNs beyond translational invariance. Hybrid pooling methods \cite{lu2015deep,lee2016generalizing} combine different types of pooling together into the a network. Stochastic pooling \cite{zeiler2013stochastic,zhai2017s3pool} randomly picks the activation in each pooling region obeying a multinomial distribution. 

Among variants of pooling, the one most close to this work is the spatial pyramid pooling (SPP) \cite{he2014spatial,he2015spatial}. SPP employs multiple pooling filters followed by concatenation, down-sampling the 2-D feature maps into a fixed-length vector, where the number of pooling filters is fixed and the size of each filter is adapted to the image size. Our multi-kernel pooling is similar to SPP in the manner of fusion of multi-scale pooling regions, while, differing from it mainly in two aspects: 1) SPP is proposed for the use of an alternative to the image cropping and warping operation. Differently, multi-kernel pooling and stacked pooling are proposed towards a boost of scale invariance of CNNs; 2) SPP is often adopted at the top of convolutional layers for the generation of a fixed-length vector for subsequent fully-connected layers. Our proposed pooling layers are more general and can substitute the vanilla pooling layers in any CNNs, especially the fully convolutional networks (FCNs) which are the state-of-the-art backbone framework for crowd segmentation, density estimation and counting.

\section{Our Approach}
We introduce very simple yet effective variants of vanilla pooling, i.e., multi-kernel pooling and stacked pooling, to improve the scale invariance of convolutional neural networks (CNNs). Please note that we take the max pooling as an example in this paper. In practice, our proposed pooling modules are totally compatible with other versions of poolings. 

\subsection{Vanilla Pooling}
The vanilla max pooling $\mathcal{P}_k$ with a kernel size of $k$ can be formalized as 
\begin{equation}
\mathcal{P}_k(z) \overset{\mathrm{def}}{=} \max_{\dot{z} \in \kappa(z,k)} X(\dot{z})
\end{equation}
where $z$ denotes a pixel position on a feature map $X \in \mathbb{R}^{W\times H}$, and $\kappa(z,k)$ denotes the square neighbourhood of $z$ with a side length of $k$. 

By applying pooling $\mathcal{P}_k$ on feature map $X$ in a manner of sliding window ($*$), we get the feature map after vanilla pooling layer
\begin{equation}
Y_{\text{vanilla}}= X * \mathcal{P}_k
\label{eq_spool}
\end{equation}

\subsection{Multi-Kernel Pooling}
In the practice of deep CNNs, a small pooling kernel, etc., $k=2$, is commonly used mainly because a larger pooling kernel may excessively discard information of the original feature map. However, a larger pooling kernel is able to provide a wider range of scale invariance for CNNs as illustrated in Fig. \ref{invariance_by_pooling}. Specifically in crowd counting, image regions of different scales generally present high visual similarity. Thus, in this work we exploit a set of poolings with different kernel sizes, i.e., multi-kernel pooling, to boost the scale invariance of a deep crowd counting model. 

The multi-kernel pooling enables a kernel set $K$ comprising of different pooling kernel sizes, such as $K=\{k_1,k_2,...,k_n\}$. As same as Eq. \ref{eq_spool}, we apply the $i$-th pooling kernel $\mathcal{P}_{k_i}$ on feature map $X$
\begin{equation}
Y_i = X * \mathcal{P}_{k_i}
\label{eq_ith_pool}
\end{equation}

\begin{figure}[t]
\centering
\includegraphics[width=0.9\linewidth]{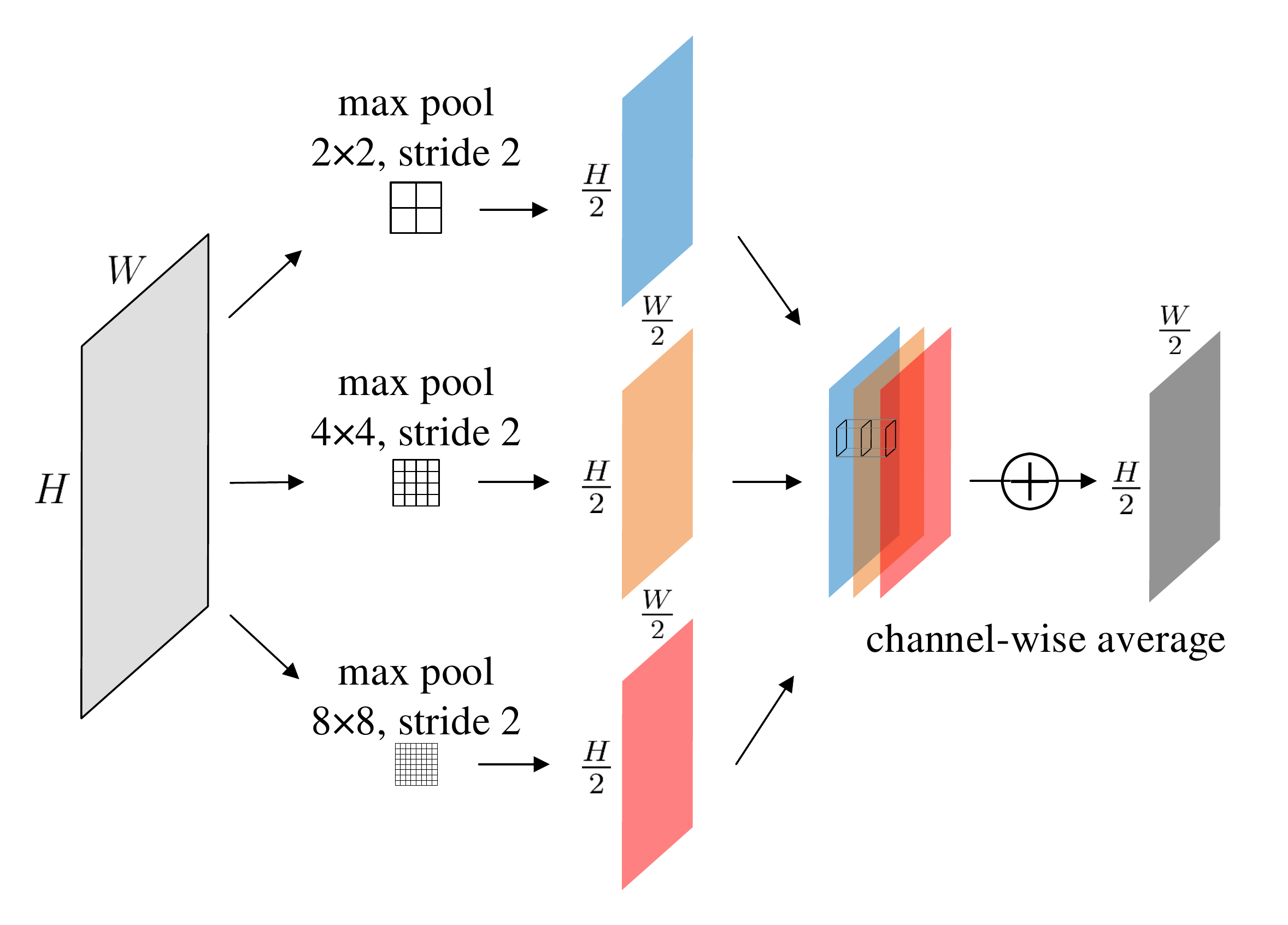}
\caption{\textbf{Multi-kernel pooling} with a set of kernels $\{2,4,8\}$ and a stride of 2. The three pooling kernels are applied on the input feature map and then concatenated with element-wise mean. } 
\label{mpool}
\end{figure}

There are many ways to concatenate the output feature maps. In this work we use \emph{element-wise mean} because: 1) It keeps the shape of original feature map; 2) It has been demonstrated to be effective in various machine learning tasks; 3) It does not introduce extra learnable parameters. Following Eq. \ref{eq_ith_pool}, the feature maps are concatenated as
\begin{equation}
Y_{\text{multi-kernel}}=\frac{1}{n} \sum_{i=1}^{n} Y_i
\label{eq_elementwise_mean}
\end{equation}

In CNNs, we often use a pooling $\mathcal{P}_k^{(s)}$ with a sliding window stride $s \geq 2$ and proper paddings to down-sample a feature map $X \in \mathbb{R}^{W\times H}$ into $\downarrow_s\hspace{-3pt} Y \in \mathbb{R}^{\frac{W}{s}\times \frac{H}{s}}$. The multi-kernel pooling with a down-sampling rate $s$ is written as
\begin{equation}
\downarrow_s \hspace{-3pt} Y_{\text{multi-kernel}} = \frac{1}{n} \sum_{k \in K} X * \mathcal{P}_k^{(s)} 
\label{eq_mpool}
\end{equation}
In theory, the multi-sized pooling kernels incorporate responses of multiple local areas into the output feature map, thus providing a wider range of scale invariance for CNNs. In addition, the fine-grained information is also preserved by those poolings with smaller kernels. Fig. \ref{mpool} illustrates an example of the multi-kernel pooling, where the kernel set $K=\{2,4,8\}$ and the stride $s=2$. In empirical studies, this configuration also shows the best performance in most cases.  

%

\subsection{Stacked Pooling}
\begin{figure}[t]
\centering
\includegraphics[width=0.98\linewidth]{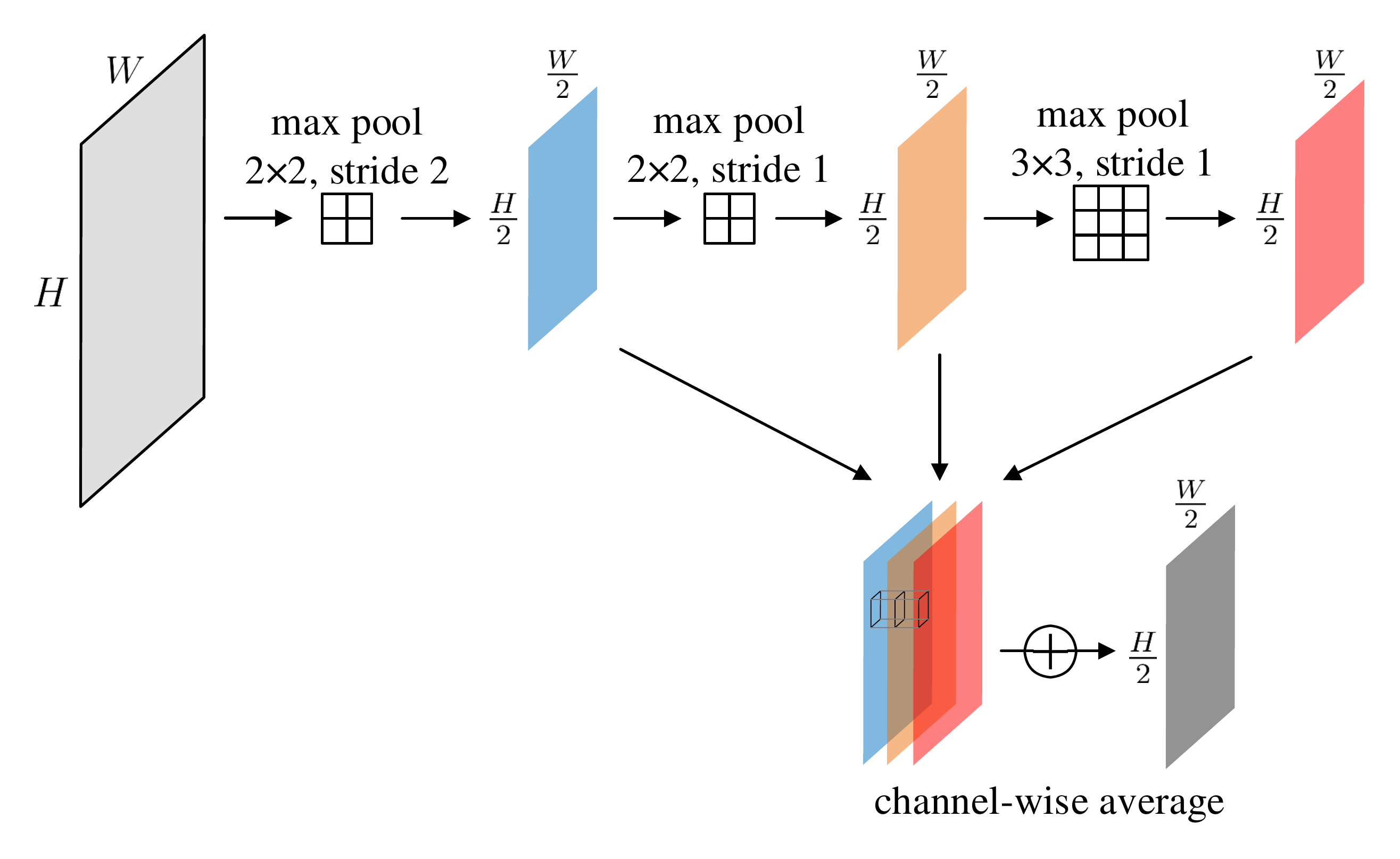}
\caption{\textbf{Stacked pooling} with a set of kernels \{2, 2, 3\}. It is an equivalent form of multi-kernel pooling shown in Fig. \ref{mpool} with less computing cost.} 
\label{stackpool}
\end{figure}

\begin{table}[t]
\small
\centering
\caption{\textbf{Time cost of pooling methods} (ms). `pool layer' is a single pooling layer. `network' is the VGG-13 network.}
\begin{tabular}{|c|c|c|c|c|}
\hline
\multicolumn{2}{|c|}{}   & vanilla  & stacked & multi-kernel \\ \hline
pool layer               & forward  & 0.11    & 0.37    & 0.84         \\ \hline
\multirow{2}{*}{network} & forward  & 6.1     & 6.6     & 7.7          \\ \cline{2-5} 
                         & backward & 13.6    & 14.1    & 15.7         \\ \hline
\end{tabular}
\label{time_cost}
\end{table}

To reduce the computing cost of multi-kernel pooling, we propose to use its equivalent form, named stacked pooling. The stacked pooling is a stack of pooling layers, where the intermediate feature maps are consecutively computed as
\begin{equation}
\downarrow_{s_i^{'}} \hspace{-3pt}Y_i^{'} = Y_{i-1}^{'} * \mathcal{P}_{k_i^{'}}^{(s_i^{'})}
\label{eq_stack_recursive}
\end{equation}
Specifically, $Y_0^{'}=X$ is the input feature map. Kernel size $k_{i}^{'}$ corresponds to $k_{i}$ with a certain transformation. Stride $s_{i=1}^{'}=s$ and $s_{i>1}^{'}=1$. Following Eq. \ref{eq_stack_recursive}, the output of stacked pooling concatenates the intermediate feature maps as
\begin{equation}
\downarrow_s \hspace{-3pt} Y_{\text{stacked}} = \frac{1}{n} \sum_{i=1}^{n} \downarrow_{s_i^{'}} \hspace{-3pt}Y_i^{'}
\label{eq_stackpool}
\end{equation}

Fig. \ref{stackpool} shows a diagram of stacked pooling which is exactly equivalent to the example of multi-kernel pooling shown in Fig. \ref{mpool}. The stacked pooling is much more efficient than multi-kernel pooling because its pooling operations are computed on down-sampled feature maps, except its first pooling kernel. Table \ref{time_cost} gives the time cost of different pooling methods w.r.t a 256$\times$256 input feature map. We can see that the stacked pooling shows a much better computing efficiency than multi-kernel pooling. On a VGG-13 network \cite{simonyan2014very}, the forward and backward time of stacked pooling is close to that of vanilla pooling, thus, ensuring its practicability.

\section{Experimental Setup}

\subsection{Datasets}
In this work, we do empirical study on two popular crowd counting datasets: ShanghaiTech \cite{zhang2016single} and WorldExpo'10 \cite{zhang2015cross}, as both the two datasets are very challenging due to diverse scene types and varying density levels.
\begin{itemize}
\item The ShanghaiTech dataset \cite{zhang2016single} consists of 1198 images
with 330,165 annotated heads. It contains two parts: Part A and Part B. Part A consists of 482 images which are randomly chosen from the Internet, having relatively larger crowd densities. Part B consists 716 images taken from the streets of metropolitan areas in Shanghai, having relatively smaller crowd densities. Part A and Part B are separately evaluated in our experiments, denoted as ShanghaiTech-A and ShanghaiTech-B.
\item The WorldExpo'10 dataset \cite{zhang2015cross} consists of 1132 annotated video sequences captured by 108 surveillance cameras. It contains a total of 199,923 annotated pedestrians in 3980 images. 
\end{itemize}
In each dataset, we randomly split the original training set into a training set and a validation set by a ratio of 9:1. We randomly crop 9 patches on each training image, where all the patches are half the size of the original image. The ground truth density map is generated by summing a 2D Gaussian kernel with a fixed $\sigma=4$ centered at every person's position \cite{lempitsky2010learning,zhang2015cross}.


\subsection{Metrics}
We use mean absolute error (MAE) and mean squared error (MSE) to evaluate the performance of different crowd counting methods:
\begin{equation}
\text{MAE} =\frac{1}{N} \sum_{i=1}^N | C_i - C^{\text{gt}}_i |, ~~~
\text{MSE}  = \sqrt{ \frac{1}{N} \sum_{i=1}^N \left( C_i - C^{\text{gt}}_i \right)^2} 
\end{equation}
\noindent
where $C_i$ is the estimated people count and $C^{\text{gt}}_i$  is the ground truth count of the $i$-th image. $N$ is the number of test images. The MAE metric indicates the accuracy of crowd estimation algorithm, while the MSE metric indicates the robustness of estimation.

\begin{table}[t]
\small
\centering
\caption{\textbf{Network architecture configurations} (shown in columns). The convolutional layer parameters are denoted as ``$\langle$kernel size$\rangle$*$\langle$kernel size$\rangle$, $\langle$channels$\rangle$''. ``pooling'' denotes a vanilla/stacked max-pooling layer. The ReLU function and the same padding operation are added after every convolutional layer. ``S'', ``M'', ``L'' represent small, medium, and large convolutional kernel size versions of base network respectively.
}
\begin{tabular}{|c|c|c|c|c|}
\hline
\multicolumn{3}{|c|}{\textbf{Base}} & \textbf{Wide} & \textbf{Deep} \\
\cline{1-3} S & M & L & & \\ 
\hline\hline
\multicolumn{5}{|c|}{input image} \\
\hline
5*5, 24 & 7*7, 20 & 9*9, 16 & 7*7, 128 & 5*5, 64 \\
&&&& 5*5, 64 \\
\hline
\multicolumn{5}{|c|}{pooling} \\
\hline
3*3, 48 & 5*5, 40 & 7*7, 32 & 5*5, 256 & 5*5, 128 \\
&&&& 5*5, 128 \\
\hline
\multicolumn{5}{|c|}{pooling} \\
\hline
3*3, 24 & 5*5, 20 & 7*7, 16 & 5*5, 128 & 3*3, 256 \\
3*3, 12 & 5*5, 10 & 7*7, 8 & 5*5, 64 & 3*3, 256 \\
&&&& pooling \\
\hline
1*1, 1 & 1*1, 1 & 1*1, 1 & 1*1, 1&  3*3, 128 \\
&&&& 3*3, 64 \\
&&&& 3*3, 32 \\
&&&& 3*3, 16 \\
&&&& 1*1, 1 \\
\hline
\end{tabular}
\label{network architecture}
\end{table}

\begin{figure*}[t]
\centering
\includegraphics[width=0.7\linewidth]{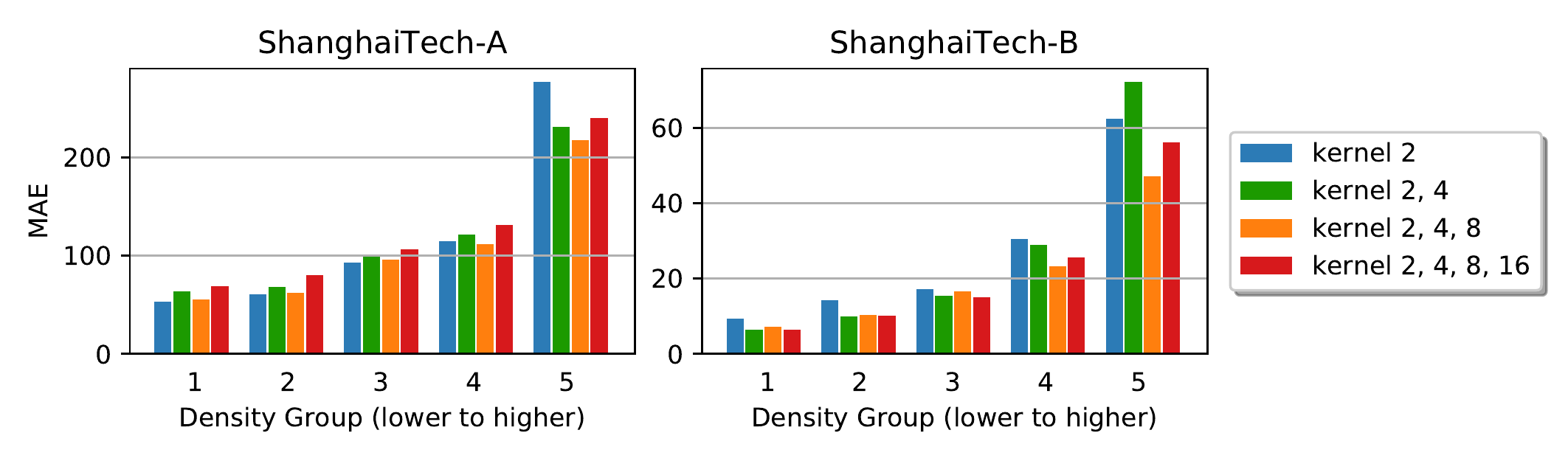}
\caption{\textbf{Experiments on kernel sizes of poolings.} The MAE, vs. the density groups from lower density to higher density.} 
\label{experiment_kernel_set}
\end{figure*}

\begin{table*}[t]
\small
\centering
\caption{\textbf{Comparison of vanilla pooling and stacked pooling on ShanghaiTech dataset.} 
}
\begin{tabular}{c|cc|cc|cc|cc|cc}
\toprule
&\multicolumn{2}{c|}{Base-S} & \multicolumn{2}{c|}{Base-M} & \multicolumn{2}{c|}{Base-L}
 & \multicolumn{2}{c|}{Wide} & \multicolumn{2}{c}{Deep} \\
\cmidrule(lr){2-3} \cmidrule(lr){4-5} \cmidrule(lr){6-7} \cmidrule(lr){8-9} \cmidrule(lr){10-11}  & MAE & MSE  & MAE & MSE & MAE & MSE & MAE & MSE & MAE & MSE\\
\midrule
\multicolumn{11}{c}{ShanghaiTech-A} \\
\midrule
vanilla & 138.56 & 220.40 & 128.91 & 200.24 & 117.14 & 181.23 & 123.11 &201.32  &99.19 & 158.88\\
stacked & \textbf{109.41} & \textbf{168.49} & \textbf{119.48} & \textbf{184.73} &\textbf{115.35}&\textbf{176.43} &\textbf{113.71} &\textbf{181.52}  &\textbf{93.98} & \textbf{150.59}\\
\midrule
\multicolumn{11}{c}{ShanghaiTech-B} \\
\midrule
vanilla & 31.54	&55.04 &25.96	&47.77 & 23.55 &42.91 &31.79 &57.22 & 20.34&37.32 \\
stacked & \textbf{23.60}&	\textbf{43.90} &\textbf{22.91}	&\textbf{39.91} &\textbf{20.96} &\textbf{37.78} &\textbf{25.84}&\textbf{47.36}& \textbf{18.02}	& \textbf{35.64} \\
\bottomrule
\end{tabular}
\label{experiment_shanghaitech}
\end{table*}

\begin{table*}[t]
\small
\centering
\caption{\textbf{Comparison of vanilla pooling and stacked pooling on WorldExpo'10 dataset.} We show the MAE performances on five test scenes.}
\begin{tabular}{l|c|c|c|c|c|c}
\toprule
& Scene \#1 & Scene \#2 & Scene \#3 & Scene \#4 & Scene \#5 & Average \\ \midrule
Wide + vanilla & 5.01 & \textbf{18.96} & \textbf{14.76} & 21.36 & 14.57 & 14.95\\
Wide + stacked & \textbf{4.72} & 22.62 & 19.85 & \textbf{14.21} & \textbf{8.43} & \textbf{13.98} \\ \midrule
Deep + vanilla & 4.08 & 18.74 & 20.68 & \textbf{23.28} & 6.84  & 14.74\\
Deep + stacked & \textbf{3.26} & \textbf{12.39} & \textbf{13.97} & 31.41 & \textbf{3.50} & \textbf{12.92} \\
\bottomrule
\end{tabular}
\label{experiment_worldexpo2010}
\end{table*}

\subsection{Network Architectures}
We evaluate our proposed stacked pooling module\footnote{Unless otherwise specified, we do experiments on stacked pooling as it is numerically equivalent to multi-kernel pooling with better efficiency.} on different backbone CNNs as shown in Table \ref{network architecture}. We exploit three types of network architectures, i.e., Base-Net, Wide-Net, and Deep-Net. The Base-Net is relatively small and it has three variants, namely ``S'', ``M'', and ``L'', coming from the three columns of Multi-Column CNN \cite{zhang2016single} and having different convolutional kernel sizes. The Wide-Net widen the Base-M Net by using more channels of feature maps. The Deep-Net follows the well-known VGG-13 network \cite{simonyan2014very} with slight modifications. We use CNNs of diverse depths, widths, and convolutional kernel sizes for a comprehensive evaluation of our method. 

\subsection{Learning Settings}
In this work, the CNNs are implemented based on PyTorch framework \cite{paszke2017automatic}. For a fair comparison, we adopt identical learning settings for vanilla pooling and stacked pooling. The Base-Net, Wide-Net, and Deep-Net are trained by an Adam optimizer \cite{kingma2014adam}. The batch size is set as 1 on ShanghaiTech dataset and set as 32 on WorldExpo'10 dataset to ensure a comprehensive evaluation with respect to batch size. The training process runs for 500 epochs on the training set. We evaluate the checkpoints on the validation set at an interval of 2 epochs. The model with the best MAE is selected as the best model used for testing. Please refer to our code\footnote{https://github.com/siyuhuang/crowdcount-stackpool} for more implementation details. 

\begin{figure*}[!htb]
\centering
\includegraphics[width=0.7\linewidth]{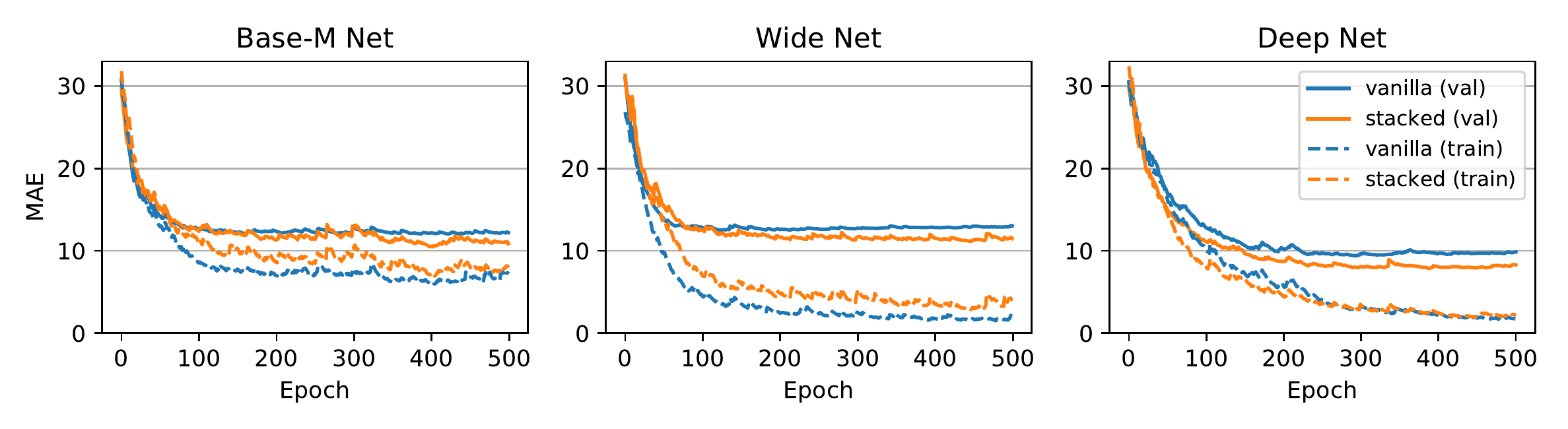}
\caption{\textbf{Learning curves.} The MAE on training and validation sets, vs. the number of training epochs.} 

\label{experiment_learning_curve}
\end{figure*}

\section{Results}
\subsection{Study on Pooling Kernels}

We first empirically study the configuration of pooling kernel set $K$ as shown in Fig. \ref{experiment_kernel_set}. The experiments are conducted on ShanghaiTech dataset and Base-M Net. Four different kernel sets, including the vanilla pooling kernel $\{2\}$ and the multi-kernel pooling kernel sets $\{2, 4\}$, $\{2, 4, 8\}$, $\{2, 4, 8, 16\}$, are evaluated. We group the test images according to the crowd densities and show the MAE of density groups from lower density to higher density.

Fig. \ref{experiment_kernel_set} shows that the vanilla pooling performs worse than our multi-kernel pooling on the high density group of ShanghaiTech-A dataset and also worse on the entire ShanghaiTech-B dataset. Among the multi-kernel pooling kernel sets, set $\{2, 4, 8\}$ performs the best with robustness on all density levels. Therefore, we employ kernel set $K=\{2, 2, 3\}$ as the default configuration of stacked pooling in the following experiments. 

\subsection{Vanilla Pooling vs. Stacked Pooling}

\paragraph{ShanghaiTech dataset} We quantitatively compare vanilla pooling based and stacked pooling by adopting them in five different CNN architectures described in Table \ref{network architecture}. Kernel set $K=\{2, 2, 3\}$ is used for stacked pooling. 

Table \ref{experiment_shanghaitech} shows the empirical results on ShanghaiTech-A and B, respectively. stacked  pooling obviously outperforms vanilla pooling by showing a superior performance over all settings of datasets, network architectures, and metrics. In regard to datasets, A part and B part of ShanghaiTech dataset vary largely with crowd densities, scenes, and camera perspectives. In regard to network architectures, the five evaluated network cover the common-used CNN architectures, from small to large, and from shallow to deep. In regard to evaluation metrics, MAE reveals the estimation accuracy of model, and MSE reveals the robustness of model. The evidences of improvements over these settings indicate that our stacked pooling module is a universally effective variant of vanilla pooling module for crowd counting task. 

The performance of pooling module on Deep-Net is what we most care, because a deep network is  generally effective and is the most often used in practical crowd counting applications. Table \ref{experiment_shanghaitech} shows that the Deep-Net is empirically better than Wide-Net and Base-Nets on ShanghaiTech dataset. In this work, we down-sample the feature maps in Deep-Net by three max pooling layers. Experimental results show that the Deep-Net is 5.2\% and 11.4\% better under MAE by adopting stacked pooling instead of vanilla pooling. In theory, the stacked pooling does not introduce extra model parameters, meanwhile, preserving more information during the down-sampling process, thus benefiting the information flow in deep layers.

\paragraph{WorldExpo'10 dataset} Table \ref{experiment_worldexpo2010} compares pooling modules on WorldExpo'10 dataset. MAE results on five different test scenes are shown respectively. We evaluate the Wide-Net and the Deep-Net for they are more often used in practice. In this experiment, the Deep-Net still performs better than the Wide-Net w.r.t. the average MAE. The MAEs across different scenes are quite different due to diverse crowd densities of the scenes. 

The stacked pooling performs better than the vanilla pooling w.r.t. the average performance and most of the testing scenes. The stacked pooling performs well on low-density scenes while showing similar performance with vanilla pooling on high-density scenes, indicating that the stacked pooling is as a whole better than the vanilla pooling for crowd images with diverse densities and various scenes. 

\paragraph{Learning curves} We investigate the training procedure of different pooling modules by studying their learning curves. Fig. \ref{experiment_learning_curve} shows the training and validation MAEs of trained models at every epoch, where the learning curves of Base-M Net, Wide-Net, and Deep-Net are shown from left to right, respectively. For better viewing, we smooth the learning curves by applying an exponential moving average (EMA) with a smoothing factor $\alpha=0.1$.

On the training set, the stacked pooling based models show higher MAEs compared to the vanilla pooling based models, where the learning curves of Base-M Net and Wide-Net distinctly show this result. In machine learning, model performance on training set generally denotes the fitting degree of a model and the training set. The vanilla pooling shows better performance on training set and worse performance on testing set, indicating that it has a better fitting capability, while, a worse generalization capability, such that it may be easier to get overfitting. The MAEs of Deep-Net with the two pooling modules are close to each other after training to convergence, mainly because the Deep-Net model is deeper and larger with more parameters, enabling a better fitting capability. In conjunction with Deep-Net, the stacked pooling also shows a good generalization performance, demonstrating its practicability in real world crowd counting scenarios.


\subsection{Study on Scale Invariance}
In previous sections, we discuss the scale invariance of CNN models, and, believe the pooling layer is one of its most important supporters. In this section, we further take some insight into the scale invariance driven by pooling modules. Specifically, we evaluate the variation ratio of feature maps after a pooling layer v.s. the scale variation of an input image. The variation ratio $\gamma$ is formulated as
\begin{equation}
\gamma=\frac{1}{|\mathcal{X}|}\sum \limits_{X \in \mathcal{X}} \frac{\sum |\hat X_{wh}-X_{wh}|} {\sum |X_{wh}|} 
\end{equation}
$X$ is a feature map within the feature maps $\mathcal{X}$ of a CNN model given an input image. We resize the input image according to a certain scaling factor $\beta$ and again calculate the corresponding feature map followed by resizing the feature map to the same size of $X$. $|\mathcal{X}|$ is the number of feature map channels. The variation ratio $\gamma$ is used to evaluate the scale invariance of a CNN model, where a CNN model with a stronger scale-invariant representation will have smaller $\gamma$ when facing the same input image of different scales.

\begin{figure}[t]
\centering
\includegraphics[width=1\linewidth]{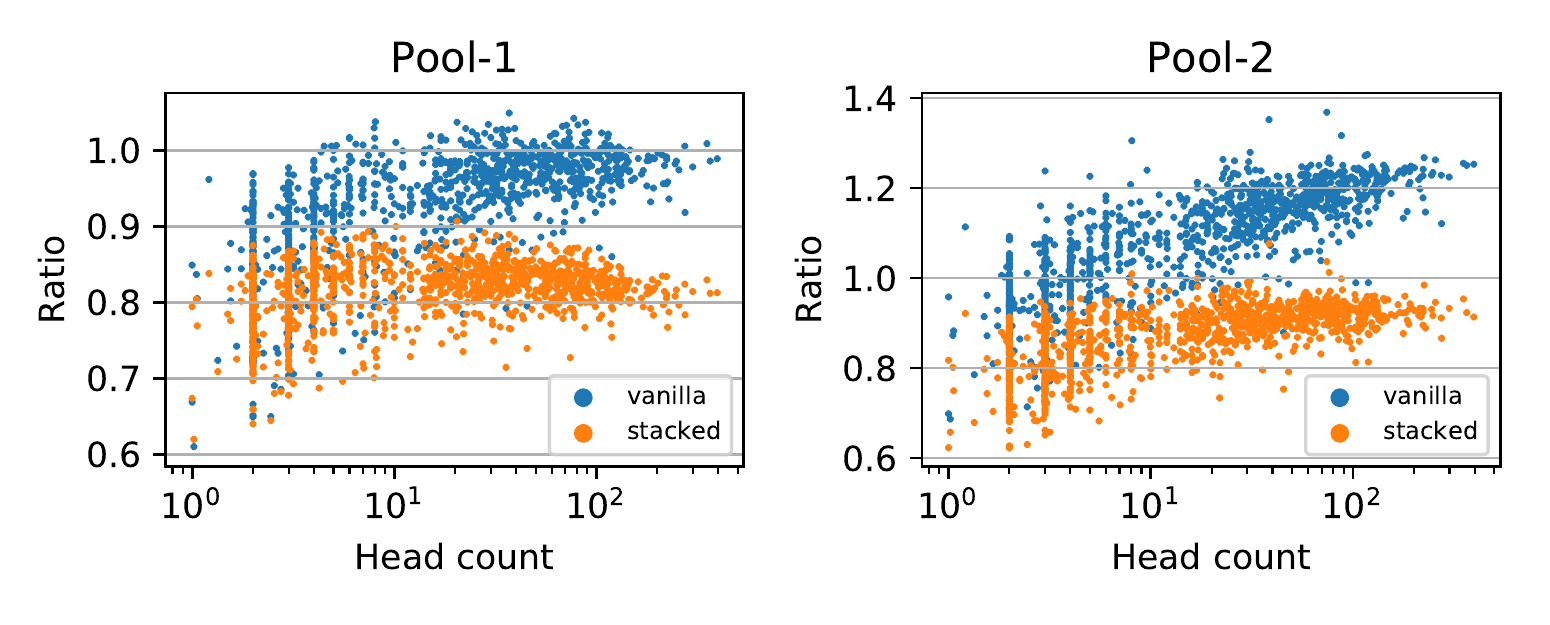}
\caption{\textbf{The scale invariance of poolings.} The variation ratio of feature maps, vs. the number of head counts. } 
\label{experiment_si}
\end{figure}

We conduct this experiment by applying Base-M Net to ShanghaiTech-B dataset, where the network is previously trained on the training set and evaluated on the testing set. Fig. \ref{experiment_si} shows the variation ratio $\gamma$ of feature maps after two respective pooling layers, given the images in the testing set. An up-sample scaling factor $\beta=2$ is adopted in this experiment. Large data points ($\gamma>2$) are ignored as outliers. 

In Fig. \ref{experiment_si}, it is distinct that the stacked pooling has a smaller variation ratio $\gamma$ than the vanilla pooling w.r.t. both pooling layers. It indicates that given the same image of different scales, the stacked pooling layer is able to provide more scale-invariant feature maps for subsequent convolutional layers, i.e., the feature maps are more consistent with the original feature maps. Such scale-invariant representation improves the generalization capability of a CNN model, especially for crowd counting datasets which have high intra-image and inter-image visual similarities. 

It is noticeable that in Fig. \ref{experiment_si} the variation ratios $\gamma$ of two pooling modules are closer on low-density images while exhibiting greater differences on high-density images. The stacked pooling has much smaller $\gamma$ than vanilla pooling on high-density images. It indicates that the stacked pooling works particularly well at high-density crowd counting cases. Fig. \ref{experiment_kernel_set} also presents this result, where the kernel set $K=\{2,4,8\}$ performs much better than a single kernel $K=\{2\}$ on high-density images.

\section{Conclusion}
In this work, we have explored an important feature in crowd counting scenario, i.e., cross-scale visual similarity, to highlight the importance of scale invariance of crowd counting models. Further, we have proposed multi-kernel pooling and stacked pooling to boost the scale invariance of CNNs, where a larger pooling range enables a stronger invariance for significant scale variation in crowd images. The stacked pooling layer is efficient and easy to implement, showing better performance than vanilla pooling layer in most cases on benchmark crowd counting datasets.

\small
\bibliographystyle{aaai}
\bibliography{paper}

\end{document}